\documentclass[preprint, 3p, twocolumn, authoryear]{elsarticle}

\makeatletter
\def\ps@pprintTitle{%
   \let\@oddhead\@empty
   \let\@evenhead\@empty
   \let\@oddfoot\@empty
   \let\@evenfoot\@oddfoot
}
\makeatother

\usepackage{standalone}

\usepackage[utf8]{inputenc}
\usepackage[english]{babel}
\usepackage[T1]{fontenc} 	
\usepackage{amsmath}
\usepackage{amssymb}
\usepackage{multirow}
\usepackage{mathptmx}
\usepackage{natbib}
\usepackage{floatrow}
\usepackage{float}
\usepackage{subcaption}
\usepackage{gensymb}
\usepackage{csquotes}
\usepackage{xcolor}

\usepackage{tikz}
\usepackage{pgfplots}
\pgfplotsset{compat=1.3}
\usepackage{tikzscale}

\usetikzlibrary{shapes}
\usetikzlibrary{plotmarks}
\newcommand\marksymbol[2]{%
\tikz[#2,scale=1.0,baseline=-0.67ex]\pgfuseplotmark{#1};%
}
\definecolor{black}{RGB}{0,0,0}
\newcommand\aucMarker{\marksymbol{*}{black}}
\newcommand\logscoreMarker{\marksymbol{square}{black}}
\newcommand\audmanMarker{\marksymbol{triangle*}{black}}

\usepackage[colorlinks=true]{hyperref}

\biboptions{sort&compress}

\newcommand\bcolorbox[2]{\bcbaux{#1}#2 \endbcb}
\def\bcbaux#1#2 #3\endbcb{%
  \colorbox{#1}{\strut#2}%
  \ifx\relax#3\relax\def\next{}\else%
    \colorbox{#1}{ \strut}%
    \allowbreak%
    \def\next{\bcbaux{#1}#3\endbcb}%
  \fi%
  \next%
}

\addto\captionsenglish{}

\journal{Expert Systems with Applications}

\newcommand\ReLU{ReLU}

\usepackage[font=footnotesize,labelfont=bf]{caption}

\begin{document}
\begin{frontmatter}

\title{Predicting Distresses using Deep Learning of Text Segments in Annual Reports}

\author[dnb] {Rastin Matin\corref{cor1}}
\ead{rma@nationalbanken.dk}
\author[diku] {Casper Hansen}
\ead{c.hansen@di.ku.dk, +4542802347}
\author[diku] {Christian Hansen}
\ead{chrh@di.ku.dk, +4542482347}
\author[dnb] {Pia M\o lgaard}
\ead{pim@nationalbanken.dk}

\cortext[cor1]{Corresponding author}

\address[dnb] {Danmarks Nationalbank, DK-1093 Copenhagen K, Denmark}
\address[diku] {Department of Computer Science, University of Copenhagen, DK-2100 Copenhagen \O, Denmark}

\begin{abstract}
Corporate distress models typically only employ the numerical financial variables in the firms' annual reports. We develop a model that employs the unstructured textual data in the reports as well, namely the auditors' reports and managements' statements. Our model consists of a convolutional recurrent neural network which, when concatenated with the numerical financial variables, learns a descriptive representation of the text that is suited for corporate distress prediction. We find that the unstructured data provides a statistically significant enhancement of the distress prediction performance, in particular for large firms where accurate predictions are of the utmost importance. Furthermore, we find that auditors' reports are more informative than managements' statements and that a joint model including both managements' statements and auditors' reports displays no enhancement relative to a model including only auditors' reports. Our model demonstrates a direct improvement over existing state-of-the-art models.
\end{abstract}


\end{frontmatter}

\section{Introduction}\label{sec:introduction}

Statistical corporate distress prediction is a binary classification task that was pioneered by \citet{Altman1968} and \citet{Ohlson1980} among others. They used a limited number of financial ratios as input and employed simplistic models such as linear discriminant analysis and logistic regression for the classification, where the financial ratios enter the model in a linear combination. Since then a range of advanced statistical methods (``machine learning'') have been applied to the problem such as gradient boosting (e.g. \cite{Caruana06}) and neural networks (e.g. \cite{Atiya2001, Tsai2008UsingNN}) including convolutional neural networks (\cite{HOSAKA2019287}). Traditionally, distress models have only employed the numerical financial variables of the firms' annual reports, i.e. structured data. However, annual reports also contain unstructured data in the form of text segments (auditors' reports and managements' statements), which is potentially a rich source of information for distress prediction. 

Since 2013 Danish regulators have required firms to provide annual reports in the open data standard for financial reporting known as eXtensible Business Reporting Language (XBRL) from which these two text segments can be easily extracted. Motivated by recent advances within natural language processing, we propose a deep learning approach for predicting corporate distresses that incorporates these text segments in addition to numerical financial variables. Using annual reports of corporate firms in Denmark from 2013 to 2016, corresponding to a total of 278\,047 firm years, our tests reveal that the auditors' reports, and to a lesser extent the managements' statements, increase the prediction accuracy compared to common state-of-the-art baseline classifiers that are based solely on structured data. This demonstrates that the unstructured data contains a signal that can enhance corporate distress prediction models. The readily availability of the data makes this study particularly valuable as current state-of-the-art can be augmented straightforwardly. 

We investigate a model employing auditors' reports, a model employing managements' statement, and a model employing both auditors' reports and managements' statements. For each of the three models, we first apply standard preprocessing techniques to the text followed by pattern extraction and recognition by using a convolutional recurrent neural network. The output from the convolutional recurrent neural network is then concatenated with numerical financial variables and the final model is estimated using two fully-connected layers. Our models further utilize an attention mechanism, which increases the model interpretability by being able to highlight words that are important for the final prediction. 

We compare performance of these three models to three competitive distress prediction models based solely on the structured data: A logistic regression, gradient boosted trees, and a neural network with the same architecture as the network that employs text. The models employing text outperform all other models. Specifically, we find that including the auditors' reports, managements' statements, and both text segments in the neural network improves prediction accuracy measured by AUC by 1.9, 1.1, and 1.8 percentage points, respectively. The performance of the model including auditors' reports is significantly better than that of the model including managements' statements, demonstrating that the auditors' reports are more informative. Including both text segments yields the same results as including only auditors' reports, illustrating that, in our sample, managements' statements do not contain information useful for distress predictions beyond what is already contained in the auditors' reports. Finally, we run the same analysis on a subsample of large firms which comprise 95\% of the debt in the economy, and find even stronger model enhancements when including auditors' reports. Given that the test is done on Danish data, and that Denmark is a relatively small economy, we believe that the gain from our textual analysis should be viewed as a lower bound relative to other larger economies, where greater amounts of data allow for improved model training, especially for data hungry models such as neural networks.

In the following section we review related works. The data and methods are described in Sections \ref{sec:data} and \ref{sec:methodology}, respectively, and in Section \ref{sec:experiment} we demonstrate the applicability of our method in predicting corporate distresses. In Section \ref{sec:case_study_ENG} we illustrate heat maps of selected word blocks, and we draw conclusions and outline future work in Section \ref{sec:conclusion}.

\section{Literature Review}\label{sec:related-works}

Traditionally, textual analysis in financial research has consisted of simple semantic analysis based on word counts (see \cite{LoughranMcDonald2011} and references herein). A recent example of this is \cite{BuehlmaierWhited2018} who use a na\"{i}ve Bayes algorithm to model the probability of firms being financially constrained by using the word count in each management's statement as input.

A small string of literature most related to our work is dedicated to textual analysis in corporate distress prediction. \citet{HajekOlej2013} categorize annual reports into six different semantic categories based on specific words found in the text. They then show, using a variety of models, that sentiment indicators improve the models' ability to predict corporate distress. \citet{RONNQVIST201757} develop a deep learning model to analyze financial news with the aim of identifying financial institutions in distress, and \citet{2017arXiv170609627C} generalize the model to include numerical financial variables as well. 

We add to the work of \citet{HajekOlej2013} by applying a highly data-driven methodology for text processing based on deep learning, thereby allowing us to learn a deeper representation of the text and extract a stronger signal. Furthermore, we provide insight to which specific text segments of the annual reports contain information most relevant for distress prediction by examining auditors' reports and managements' statements separately. This data-driven methodology for textual analysis is close to that of \citet{RONNQVIST201757} and \citet{2017arXiv170609627C}. However, we learn the textual representation end-to-end, compared to \citet{2017arXiv170609627C} who first learn a representation of the text, unrelated to the specific task, and then use it together with numerical financial variables. Our approach enables the textual representation to look for signals in the reports which are important for the task of distress prediction. Furthermore, we base our analysis on annual reports which are homogeneous across firms, whereas news articles tend to focus on specific stories which the public finds interesting.

More thoroughly studied is the concept of distress modelling using neural networks and other machine learning techniques based solely on numerical financial variables (see e.g. \citet{ Jones2017, ZIEBA201693, sun2014predicting, sun2017dynamic}). The existing literature tends to find that tree-based algorithms, i.e. random forest and gradient boosted trees, outperform neural networks when only numerical financial variables are included in the models. Hence, we benchmark our model against not only a neural network, but also state-of-the-art gradient boosted trees in addition to a more traditional logistic regression model.

\section{Data}\label{sec:data}
Our data set is based on the data used in \citet{PD_WP}. It consists of non-consolidated annual reports filed by all Danish non-financial and non-holding private limited and stock-based firms. This data is augmented with firm characteristics such as age, sector, and legal status from the Danish Central Business Register. In total the data set consists of the 50 numerical financial variables (44 continuous and 6 categorical) listed in Table \ref{table:covariates}. The list follows from application of a thresholded Lasso and the numerical variables are winsorized at 5\% and 95\% quantiles for enhanced performance (\citet{PD_WP}).

We further include the auditors' reports and managements' statements found within the very same annual reports. The management's statement describes the management's opinion on the given fiscal year and its outlook on the firm's future. The auditor's report consists of several paragraphs, where the (presumably) most important for distress prediction contains the auditor's opinion of the annual report and summarizes the financial health of the firm. In this section the auditor will explicitly state any concerns regarding the continued operation of the firm or any disagreements with the management's statement. We include all available paragraphs of these two text segments in our model. 

We formally seek to model the probability of a given firm entering into distress, where \enquote{distress} refers to \enquote{in bankruptcy}, \enquote{bankrupt}, \enquote{in compulsory dissolution}, or \enquote{ceased to exist following compulsory dissolution}. Firms that cease to exist due to other reasons and firms that enter into distress more than two years after the last annual report is made public are excluded from our sample.

Our sample period starts in 2013, which marks the point in time where statements are available in the XBRL-format\footnote{Extracted data is delivered to us by Bisnode.}. The sample ends in 2016, and marks the last year where we can observe realized distresses. As of 2006 small and newly established firms in Denmark are not required by law to include an auditor's report in their annual report.\footnote{We refer to the Danish Commerce and Companies Agency for details.} As we want to directly compare the model when employing either or both of the two text segments, we therefore limit our data set to statements that contain both a management's statement and an auditor's report. This constraint removes 88\,343 firm years from the data set (corresponding to 24.1\%) and our final data set contains 278\,047 firm years across 112\,974 unique firms and 8\,033 distresses.\footnote{Our distress definition implies that firms can enter into distress multiple times. In our sample, 47 of the distresses are such recurrent events.} 

The 25\%, 50\% and 75\% percentiles of the managements' statements and the auditors' reports are 37, 54, and 83 words and 187, 205, and 219 words, respectively. The greater length of the auditors' reports may not necessarily imply more relevant information, as auditors' reports typically contain standardized paragraphs that describe the responsibilities of the auditor and summarize the accounting practices.

\begin{table}[htbp]
\resizebox{\textwidth}{!}{%
\centering
\caption{\textbf{Numerical financial variables and their type (continuous or categorical)}. The table lists the 50 numerical financial variables included in the models. An asterisk denotes scaling by the firm size, which is defined as the total debt of the firm when equity is negative and otherwise total assets. We refer to \cite{PD_WP} for the definition of each variable and details regarding the variable selection procedure.} \label{table:covariates}
\begin{tabular}{cl}
\hline
Type &Input variable\\
\hline
Continuous  &Accounts payable$^*$\\
			&Accounts receivable$^*$\\
			&Change in log size\\
			&Corporation tax$^*$\\
			&Current assets$^*$\\
			&Deferred tax$^*$\\
			&Depreciation$^*$\\
			&EBIT$^*$\\
			&Equity/invested capital\\
			&Equity$^*$\\
			&Expected dividends$^*$\\
			&Financial assets$^*$\\
			&Financial income$^*$\\
			&Financing costs$^*$\\
			&Fixed costs$^*$\\
			&Ind. EW avg. net profit$^*$\\
			&Interest coverage ratio\\
			&Inventory$^*$\\
			&Invested capital$^*$\\
			&Land and buildings$^*$\\
			&Liquid assets$^*$\\
			&log(age)\\
			&log(size)\\		
			&Long-term bank debt$^*$\\
			&Long-term debt$^*$\\
			&Long-term mortgage debt$^*$\\
			&Net profit$^*$\\
			&Other operating expenses$^*$\\
			&Other receivables$^*$\\
			&Other short debts$^*$\\
			&Personnel costs$^*$\\
			&Prepayments$^*$\\
			&Provisions$^*$\\
			&Quick ratio\\
			&Receivables from related parties$^*$\\
			&Relative debt change\\
			&Retained earnings$^*$\\
			&Return on equity (\%)\\
			&Short-term bank debt$^*$\\
			&Short-term mortgage debt$^*$\\
			&Tangible fixed assets$^*$\\
			&Tax expenses$^*$\\
			&Total receivables$^*$\\			
\hline
Categorical &Has prior distress\\
			&Is private limited (Danish ``Anpartsselskab'')\\
			&Large debt change\\
			&Negative equity\\
			&Region\\
			&Sector\\
\hline
\end{tabular}}
\end{table}

\subsection{Text Preprocessing}\label{sec:preprocess}
To preprocess the unstructured data we apply the following five steps to the auditor's report and management's statement of each annual report:
\begin{enumerate}
    \item Remove punctuation marks, newlines and tabs and convert to lowercase.
    \item Apply the Porter stemming algorithm \citep{PorterStemmer} with the NLTK library \citep{BirdKleinLoper09} to obtain the word stems and enable words to be evaluated in their canonical forms.
    \item Remove stop words including numbers (i.e. dates and amounts of money) in order to avoid overfitting the network to a particular format. Numbers are replaced by a generic number token.
    \item Perform named-entity recognition using spaCy \citep{spacy2} in order to strip the text of any names and entities that may lead to overfitting in the training process and reduce generalizability.
    \item During construction of the vocabulary we ignore words that have less than 25 occurrences across the entire data set.
\end{enumerate}
Steps 1, 2, and 3 are considered standard procedures with the purpose of reducing unique tokens in the text in order to reduce the variability across the reports. The purpose of steps 4 and 5 is to create a model that generalizes well by removing all names and entities from the texts. The aim is to prevent the model from overfitting to certain characteristics such as firm names, auditor names, and locations. The pruning of low frequency words (step 5) is done explicitly as Danish word embeddings are trained on only a dump of the Danish Wikipedia, and rare words are therefore not represented well.


\section{Models for Corporate Distress Prediction}\label{sec:methodology}
In this section we first describe our network architecture for predicting corporate distresses, which incorporates either or both of the two text segments in addition to numerical financial variables, followed by an overview of the competitive baseline models used for comparison in the experimental evaluation. 

\subsection{Main Model}
We first provide an overview of our model in order to improve the understanding of its individual parts:
\begin{description}
    \item[Word Representation:] We use word embeddings to map each word of a text segment into a dense vector in a feature space, where semantically similar words are close to each other. Using this we split the given text segment into half-overlapping blocks of words.
    \item[Pattern Extraction:] Using the embedded word blocks we utilize a convolutional neural network (CNN) to extract patterns from each block and learn a lower dimensional representation.
    \item[Pattern Understanding:] The pattern output from the CNN is fed to a recurrent neural network (RNN), and the final text representation is calculated as an attention-weighted sum of the individual RNN outputs.
    \item[Feature Extensions and Prediction:] We concatenate the attention-weighted sum with the numerical variables listed in Table \ref{table:covariates} and feed it through two fully-connected layers to arrive at the final corporate distress probability prediction.
\end{description}
In the following we will explain in detail the individual parts, and we refer to Figure \ref{fig:network_architecture} for a visual description of the network architecture.

\subsubsection{Word Representation}
We choose to represent the semantics of each word through state-of-the-art word embeddings, which is a mapping from a word to a dense vector representation, where semantically similar words are close to each other. We use the \textit{word2vec} model \citep{mikolov2013distributed}, specifically the skip-gram model. The objective of the skip-gram model is for a word to be able to predict its surrounding words. For a sequence of words $w_1, w_2, ..., w_n$ we maximize the log probability $p$
\begin{align}
    \max \frac{1}{n} \sum_{t=1}^n \; \sum_{j=-c, j \neq 0}^{c} \log p(w_{t+j} | w_t)
\end{align}
where $c$ denotes the number of words before and after the current word to consider, which is fixed to 5 in this paper. Negative sampling is used to compute $\log p$ and the words are sub-sampled proportional to their inverse frequency. In \textit{word2vec} semantically similar words have a high cosine similarity between them and allows for vector calculation of words such that e.g. $\textbf{king} - \textbf{man} + \textbf{woman}$ is very close to $\textbf{queen}$. We do not learn the word embedding from scratch, but rather exploit a model\footnote{\url{https://github.com/Kyubyong/wordvectors}} pre-trained on a dump from the Danish Wikipedia, and fine-tune it during network training. 

To prepare a given text segment of an annual report for the CNN in the next step, we create half-overlapping blocks of words with a step size of $k$, such that the first block consists of word $w_1, w_2, ..., w_k$ and the next block of $w_{k/2}, w_{k/2+1}, ..., w_{k/2 + k}$. If the word embedding maps to vectors in $\mathbb{R}^v$, then each of these blocks $B$ are a matrix of size $k \times v$, where $v=300$ in our setup. We consider these embedded word blocks the input to our model, and the CNN in the next step will extract patterns from these.

\subsubsection{Pattern Extraction}
For each block $B$ we apply a single-layer CNN consisting of a convolution and a max-pooling step. The purpose of the convolution is to extract matching patterns between learned filters and the embedded word block in order to learn a representation that is able to infer which patterns are important for the distress prediction task. We learn $m$ filters from a block $B$, where each filter generates a new representation $\mathbf x$, i.e. we end up with $\textbf{x}^{(p)}$ representations for $p\in \{1,2,\ldots, m\}$. The $i$th entry in $\textbf{x}^{(p)}$ is given by
\begin{align}
    \textbf{x}_{i}^{(p)} = \sum_{s=0}^{\gamma-1} \sum_{j=0}^{v-1} W_{s,j}^{(p)} B_{i+s,j}
\end{align}
where $\gamma$ denotes the number of words considered in the filter (should be less than $k$), \(v\) denotes the size of the word embedding, and $W$ is a learned parameter of size $\gamma \times v$. The filter is only applied when it is not out of bounds, resulting in $\textbf{x}^{(p)}$ being a vector of size \(k-\gamma+1\). 

Each $\textbf{x}^{(p)}$ is then max-pooled to provide a smooth signal. We denote a max-pooled vector $\textbf{x}^{(p)}$ as $\textbf{z}^{(p)}$, where the $i$th entry is given by
\begin{align}
\textbf{z}_{i}^{(p)} = \max \textbf{x}_{s}^{(p)}, \quad s &\in         \big[i,i+\tau-1\big]
\end{align}
where $\tau$ denotes the pool size. The max-pooling is only applied when it is not out of bounds, resulting in $\textbf{z}^{(p)}$ having the size \(k-\gamma-\tau+2\). Finally, the results of each filter are concatenated, yielding a final vector representation $\textbf{z}$ for each block of size \((k-\gamma-\tau+2)\times m\).

\subsubsection{Pattern Understanding}
To be able to learn the semantics and sequential nature of the text as a whole, we use an RNN on the block representations we derived in the previous section. Specifically, we employ a Long Short-Term Memory (LSTM) network \citep{lstm} on the block representations $\{\textbf{z}_1,\textbf{z}_2,...,\textbf{z}_T\}$, where $T$ refers to the number of blocks the text segment of the annual report is divided into. At a given step $t$, an LSTM cell takes three inputs: The $t$th word block representation $\textbf{z}_t$, the previous output $\textbf{h}_{t-1}$, and the previous cell state $\textbf{c}_{t-1}$. The cell then computes $\textbf{h}_t$ and $\textbf{c}_t$ by doing the following
\begin{align}
    \textbf{f}_t &= \sigma \big( W_f \cdot [\textbf{h}_{t-1},\textbf{z}_t] + \textbf{b}_f \big) \\
    \textbf{i}_t &= \sigma \big( W_i \cdot [\textbf{h}_{t-1},\textbf{z}_t] + \textbf{b}_i \big) \\
    \textbf{u}_t &= \tanh \big( W_u \cdot [\textbf{h}_{t-1},\textbf{z}_t] + \textbf{b}_u \big) \\
    \textbf{o}_t &= \sigma \big( W_o \cdot [\textbf{h}_{t-1},\textbf{z}_t] + \textbf{b}_o \big) \\
    \textbf{c}_t &= \textbf{f}_t \odot \textbf{c}_{t-1} + \textbf{i}_t \odot \textbf{u}_t \\
    \textbf{h}_t &= \textbf{o}_t \odot \tanh \textbf{c}_t
\end{align}
where $\sigma$ and $\tanh$ are element-wise sigmoid and hyperbolic tangent functions, $\odot$ is element-wise multiplication, all $W$ and $\textbf{b}$ are learned parameters, and $\textbf{f}_t$, $\textbf{i}_t$, $\textbf{o}_t$ are known as the forget, input and output gates of the LSTM cell. 

Instead of using the output at the final step $\textbf{h}_T$, we use an attention-weighted sum of the step-wise outputs. Specifically, for each $\textbf{h}_t$ we learn a scalar $score(\textbf{h}_t)$ that signifies the importance of that specific $\textbf{h}_t$. The score is computed using a single layer of size 1 with a linear activation. We use the softmax-function to normalize each scalar to derive each attention weight $\alpha_t$
\begin{align}
    \alpha_t = \frac{\exp(\textit{score}(\textbf{h}_t))}{\sum_{i=1}^T \exp(\textit{score}(\textbf{h}_i))}
\end{align}
We can then derive the final attention-weighted textual representation by the weighted sum
\begin{align}
    \textbf{h}_{\text{final}} = \sum_{t=1}^T \alpha_t \textbf{h}_t
\end{align}
The benefit of using attention is to enable the model to focus its attention on fewer, but more important, parts of the text to learn a better descriptive representation \citep{zhang2018deep}. Additionally, it enables an improved gradient flow in longer texts, such as the ones we work with in this paper.

\subsubsection{Feature Extension and Prediction}
We now have a dense textual representation, $\textbf{h}_{\text{final}}$, which we concatenate with the numerical variables $\textbf{h}_{\text{num}}$ of Table \ref{table:covariates}, yielding a vector $\textbf{h}_{\text{concat}}$ with a length equal to the sum of the number of handcrafted features and LSTM cell size. This concatenated representation is passed through two fully-connected layers of size 200 and 50 with a single neuron layer as the last step with a sigmoid activation. This is to allow the textual representation to interact with the numerical variables before doing the final prediction. The two layers of size 200 and 50 use the rectified linear unit (ReLU) activation function
\begin{align}
    \textbf{h}_{\text{concat}} &= [\textbf{h}_{\text{final}}, \textbf{h}_{\text{num}}] \\
    \textbf{l}_1 &= \ReLU \big( W_1 \cdot \textbf{h}_{\text{final}} + \textbf{b}_1 )\\
    \textbf{l}_2 &= \ReLU \big( W_2 \cdot \textbf{l}_1 + \textbf{b}_2 )\\
    PD  &= \sigma \big( W_3 \cdot \textbf{l}_2 + \textbf{b}_3 \big)
\end{align}
where $PD$ denotes the predicted distress probability. We train the network using the Adam optimizer \citep{kingma2014adam} and use the binary cross-entropy as the loss function. We will detail the parameters of the cross-validated network configuration in Section \ref{sec:tuning}. 

It is well known that neural networks are susceptible to overfitting \citep{Gu2017}. As a way of regularizing the training process we set aside 10\% of the training set as a validation. The validation set is used for early stopping, i.e. we terminate the gradient descent when the network starts to overfit.

\begin{figure*}[t]
    \centering
    \includegraphics{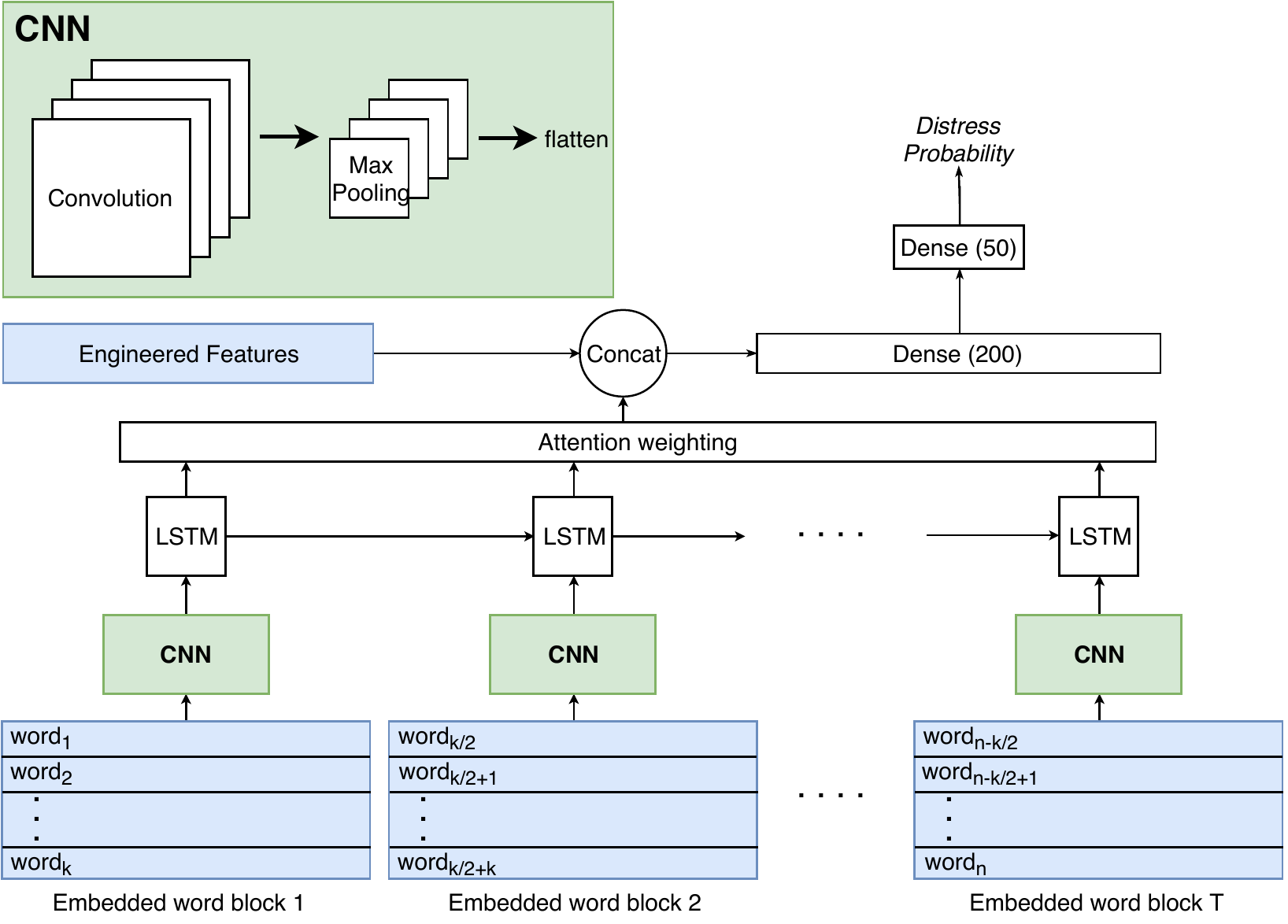}
    \caption{\textbf{Network architecture.}}
    \label{fig:network_architecture}
\end{figure*}

\subsection{Parameter Tuning in the Main Model}\label{sec:tuning}
We tune the neural network\footnote{We implemented the neural models in TensorFlow \citep{tensorflowpaper}.} using cross-validation over the hyperparameter space. For the convolutional neural network we consider block sizes in the set $\{10, 15,20\}$, number of filters in $\{40, 60\}$, and pool sizes in $\{2, 4, 6\}$. For the recurrent neural network we consider LSTM cell sizes in $\{50, 100, 150\}$. Lastly, we consider learning rates in $\{10^{-3}, 10^{-4}\}$. We run for a maximum of 10 epochs which, however, was never reached due to early stopping and use a batch size of 64 due to memory constraints. We observe that the results\footnote{AUC, described in Section \ref{sec:eval_measures}, is used as the performance metric during parameter-tuning.} are robust across this set of parameters to within one standard error for both text segments. Consequently, we use typical values in our models. For the convolutional neural network this means a block size of $k=20$ and $m=40$ filters with a pool size of $\tau=4$. The recurrent neural network uses an LSTM with a cell size of 100, and we employ a learning rate of $10^{-3}$. We set $\gamma$ (number of words to convolve over) to half the block size, i.e. $\gamma=10$. The results of the grid search are illustrated in Figure \ref{fig:gridsearch} of the Appendix for both text segments.

\subsection{Baseline Models}\label{sec:baselines}
We implement three baseline models based solely on the numerical financial variables, against which we benchmark our main model. 

First, we implement a neural network based on the same architecture as our main model, but where the textual component is not included. That is, the model consists of the two top dense layers in Figure \ref{fig:network_architecture}. This model serves as a natural benchmark as it will reveal the impact of the text segments on the prediction accuracy. 

Secondly, we implement a model based on gradient boosted trees, specifically XGBoost (XGB) \citep{chen2016xgboost}, which typically performs better than neural networks for predicting corporate bankruptcies \citep{ZIEBA201693, Jones2017}. It is an ensemble technique which recursively combines multiple relatively simple models, so-called (weak) base learners which consist of regression trees, to produce a highly accurate prediction rule. 

Finally, we implement a logistic regression (logit) which is a relatively simple, yet very common, choice for distress models (see e.g. \cite{Shumway01, Chava04, Beaver05, Campbell08}).


\section{Experimental Evaluation}\label{sec:experiment}

\subsection{Evaluation Measures}\label{sec:eval_measures}
We quantify model performance using two metrics, AUC and log score. The AUC (Area Under the receiver operating characteristics Curve) is a commonly used metric in distress prediction models. It measures the probability that a model places a higher risk on a random firm that experiences a distress event in a given year than a random firm that does not experience a distress event in a given year. Hence, 0.5 is random guessing and 1 is a perfect result. 

AUC is only a ranking measure; a model may rank the firms well, but perform poorly in terms of the level of the predicted probabilities. Generally, we are interested in well-calibrated probabilities in addition to their ranking. Thus, we look at the log score as well which takes into account the individual predicted probabilities. The log score, $\mathcal L$, for a given model is defined as
\begin{align}
\mathcal L = -\frac{1}{N}\sum_{i,t}{(y_{it}\log(\hat p_{it}) + (1-y_{it})\log(1-\hat p_{it}))}
\end{align}
where $\hat p_{it}$ is the model-predicted distress probability of firm $i$ in year $t$, $y_{it}$ is a dummy that is equal to 1 if firm $i$ enters into distress in year $t$ and 0 otherwise, and $N$ is the sample size. A smaller log score implies a better model fit. 

\subsection{Main Results}\label{sec:exp-leave-out-k-companies}
This section presents the main results of the out-of-sample tests of our models. We use 10-fold cross-validation, where the folds are constructed by sampling firms. Ideally, we would have used an expanding window of data to estimate the models and forecast the probability of the firms entering into distress two years after the estimation window closes, thereby mimicking the true forecasting situation. However, this forecasting scheme is not viable in the current study due to the limited number of years in our data set. 

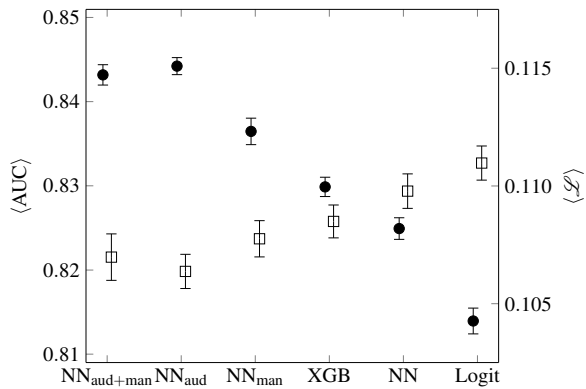
\begin{figure}[h]
\centering
\resizebox{\linewidth}{!}{
\begin{tikzpicture}

\tikzstyle{every node}=[font=\large]

\pgfplotsset{
    scale only axis,
    xmin=-2.25, xmax=7.5,
    xtick={-1.75,0,1.75,...,7},
    xticklabels={$\text{NN}_{\text{aud}+\text{man}}$, $\text{NN}_{\text{aud}}$, $\text{NN}_{\text{man}}$, XGB, $\text{NN}$, Logit}}

\begin{axis}[
  axis y line*=left,
  ytick={0.81,0.82,...,0.86},
  ymin=0.809,
  ymax=0.851,
  ylabel=$\langle \text{AUC} \rangle$,
  scatter/classes={a={}}]

    \addplot [scatter, only marks, mark=*,mark size=3pt]
    plot [error bars/.cd, y dir = both, y explicit]
    table[row sep=crcr, x index=0, y index=1, y error index=2,]{
    -1.85 0.8432110 0.00120972\\
    -0.10 0.8442520 0.001012943\\
     1.65 0.8364890 0.001574078\\
     3.40 0.8298910 0.001148603\\
     5.15 0.8249350 0.001285021\\            
     6.90 0.8139433 0.001533835\\            
    };
\end{axis}

\begin{axis}[
  axis y line*=right,
  axis x line=none,
  ymin=0.1025, ymax=0.1175,
  ytick={0.105, 0.110, 0.115},
  y tick label style={
        /pgf/number format/.cd,
            fixed,
            fixed zerofill,
            precision=3,
        /tikz/.cd
    },
  ylabel=$\langle \mathcal L \rangle$,  
  scatter/classes={a={}}]

    \addplot [scatter, only marks, mark=square,mark size=3pt]
    plot [error bars/.cd, y dir = both, y explicit]
    table[row sep=crcr, x index=0, y index=1, y error index=2,]{
    -1.65 0.1069780 0.00098668\\
     0.10 0.1063710 0.0007261733\\
     1.85 0.1077570 0.0007699207\\
     3.60 0.1084937 0.0006982064\\
     5.35 0.1097810 0.0007324107\\   
     7.10 0.1109730 0.0007248852\\        
    };
\end{axis}

\end{tikzpicture}
}
\caption{\textbf{Average AUC and log score.} The figure shows average AUC (\protect\aucMarker \,\, left axis) and average log score (\protect\logscoreMarker \,\, right axis) with one standard error band for each of the six models. Averages and standard errors are calculated based on 10 folds, which are constructed by sampling firms.}
\label{fig:out_sample_plots_cvr}
\end{figure}

The average AUC and log score across folds with one standard error bands are shown in Figure \ref{fig:out_sample_plots_cvr}, where the neural network without text is denoted $\text{NN}$ and the neural networks with text are denoted $\text{NN}_\text{aud + man}$, $\text{NN}_\text{aud}$, and $\text{NN}_\text{man}$ depending on the text segments included in the model (\emph{aud} refers to auditor's report and \emph{man} to the management's statement). This nomenclature will be used for the remainder of the paper. We observe that the neural networks with text have higher AUC and smaller log score than all baseline models. That is, the models with text are both better at ranking firms by their riskiness and provide better model fits in general. 
\\

The results of Figure \ref{fig:out_sample_plots_cvr} are furthermore summarized in Table \ref{table:CV_performance} alongside $p$-values from a paired two-tailed $t$-test comparing results of each baseline model to $\text{NN}_\text{aud + man}$, $\text{NN}_\text{aud}$, and $\text{NN}_\text{man}$, respectively. A statistically significant improvement is observed in the models with text relative to any of the baseline models, both when it comes to AUC and log score. Specifically, we find that including auditors' reports, managements' statements, and both in the neural network increases the AUC by 1.9, 1.1,  and 1.8 percentage points, respectively. That is, both the auditors' reports and the managements' statements have significant predictive power beyond what is captured by the numerical financial variables themselves. 
\\
\begin{table}[h]
\centering
\begin{subtable}{\textwidth}
\centering
\begin{tabular}{lcccc}
\hline
Model &$\langle \text{AUC}\rangle$ &$p_{\text{aud}+\text{man}}$ &$p_{\text{aud}}$ &$p_{\text{man}}$\\
\hline
$\text{NN}_{\text{aud}+\text{man}}$    &$0.843$ &--      &--       &--      \\
$\text{NN}_{\text{aud}}$               &$0.844$ &$0.233$ &--       &--      \\
$\text{NN}_{\text{man}}$               &$0.836$ &$0.000$ &$0.000$  &--      \\
\hline
XGB                                    &$0.830$ &$0.000$ &$0.000$  &$0.003$ \\
$\text{NN}$                            &$0.825$ &$0.000$ &$0.000$  &$0.000$ \\
Logit                                  &$0.814$ &$0.000$ &$0.000$  &$0.000$ \\
\hline
\end{tabular}
\caption{AUC}
\end{subtable}%
\\%
\bigskip
\begin{subtable}{\textwidth}
\centering
\begin{tabular}{lcccc}
\hline
Model &$\langle \mathcal L \rangle$ &$p_{\text{aud}+\text{man}}$ &$p_{\text{aud}}$ &$p_{\text{man}}$\\
\hline
$\text{NN}_{\text{aud}+\text{man}}$    &$0.1070$ &--       &--        &--       \\
$\text{NN}_{\text{aud}}$               &$0.1064$ &$0.4263$ &--        &--       \\
$\text{NN}_{\text{man}}$               &$0.1078$ &$0.1471$ &$0.0032$  &--       \\
\hline
XGB                                    &$0.1085$ &$0.0643$ &$0.0001$  &$0.0372$ \\
$\text{NN}$                            &$0.1098$ &$0.0005$ &$0.0000$  &$0.0001$ \\
Logit                                  &$0.1110$ &$0.0001$ &$0.0000$  &$0.0000$ \\
\hline
\end{tabular}
\caption{Log score $\mathcal L$}
\end{subtable}
\caption{\textbf{Average AUC and log score.} The table shows (a) average AUC and (b) average log score, where $p_{\text{aud}+\text{man}}$, $p_{\text{aud}}$, and $p_{\text{man}}$ denote $p$-values from a paired two-tailed $t$-test between the scores of the current model and the three models including text. Averages and standard errors are calculated based on 10 folds, which are constructed by sampling firms.}\label{table:CV_performance}
\end{table}%

The AUC and log score of $\text{NN}_\text{aud}$ is significantly better than that of $\text{NN}_\text{man}$, i.e. the auditors' reports contain more valuable information than the managements' statements. There can be several explanations for that. First, the auditors' reports are longer, enabling the neural network to learn a better representation of the text. Secondly, and more importantly, the management's statement about its own business is likely to be less objective and biased towards a brighter outlook on the future, whereas the independent auditor's report contains the auditor's unbiased professional opinion. Interestingly, there is no significant difference between $\text{NN}_\text{aud}$ and $\text{NN}_\text{aud+man}$. If anything, there is a small tendency for $\text{NN}_\text{aud}$ to perform better than $\text{NN}_\text{aud+man}$. This finding implies that, though there is information in the managements' statements which is not captured by the financial variables, all information in the managements' statements is captured by the auditors' reports. Hence, it might be preferable to focus only on the auditors' reports and leave out the managements' statements in future work.

\subsection{Results for Large Firms}
We repeat the above test, but only include firms of a size\footnote{Cf. Table \ref{table:covariates} we define firm size as the total debt of the firm when equity is negative and otherwise total assets.} greater than 5 million DKK. These firms correspond to only 35.4\% of the sample size, but 95.4\% of the total debt. It is of greater interest to quantify the performance among these dominating firms as they hold the majority of the total assets and debt in the economy. Model estimation is still done on the full sample. 

The results are summarized in Table \ref{table:CV_performance_large}, and we observe that all models yield better AUC and log score compared to the previous experiment. This is not surprising as large firms likely provide more accurate annual reports which lead to more accurate model predictions.\footnote{The large drop in log score can also in part be due to a smaller distress rate among large firms. The change in the composition of the outcome variable will by construction reduce the log score.} Interestingly, the AUC now increases by 2.6 percentage points when adding auditors' reports to the neural network, where the increase was 1.9 percentage points in the previous experiment. We speculate that this is due to the auditors' reports of the larger firms being of a higher quality and more informative, implying that the neural network can extract more information from them. On the contrary, we do not see an increased enhancement in AUC when it comes to the managements' statements, and the difference in AUC between XGB and $\text{NN}_\text{man}$ is now insignificant. This highlights that there is information to be extracted from the auditors' reports, in particular when it comes to large firms, whereas the managements' statements are less informative. The loss of significance is possibly caused by the smaller sample size, resulting in more extreme values of the individual folds. 
\\
\begin{table}[htbp]
\centering
\begin{subtable}{\textwidth}
\centering
\begin{tabular}{lcccc}
\hline
Model &$\langle \text{AUC}\rangle$ &$p_{\text{aud}+\text{man}}$ &$p_{\text{aud}}$ &$p_{\text{man}}$\\
\hline
$\text{NN}_{\text{aud}+\text{man}}$    &$0.877$ &--      &--       &--      \\
$\text{NN}_{\text{aud}}$               &$0.879$ &$0.562$ &--       &--      \\
$\text{NN}_{\text{man}}$               &$0.864$ &$0.013$ &$0.004$  &--      \\
\hline
XGB                                    &$0.860$ &$0.000$ &$0.000$  &$0.290$ \\
$\text{NN}$                            &$0.853$ &$0.000$ &$0.000$  &$0.002$ \\
Logit                                  &$0.834$ &$0.000$ &$0.000$  &$0.000$ \\
\hline
\end{tabular}
\caption{AUC}
\end{subtable}%
\\
\bigskip
\begin{subtable}{\textwidth}
\centering
\begin{tabular}{lcccc}
\hline
Model &$\langle \mathcal L \rangle$ &$p_{\text{aud}+\text{man}}$ &$p_{\text{aud}}$ &$p_{\text{man}}$\\
\hline
$\text{NN}_{\text{aud}+\text{man}}$    &$0.0611$ &--       &--       &--        \\
$\text{NN}_{\text{aud}}$               &$0.0611$ &$0.9815$ &--       &--        \\
$\text{NN}_{\text{man}}$               &$0.0627$ &$0.0551$ &$0.0095$ &--        \\
\hline
XGB                                    &$0.0629$ &$0.0085$ &$0.0127$  &$0.6588$ \\
$\text{NN}$                            &$0.0640$ &$0.0036$ &$0.0001$  &$0.0046$ \\
Logit                                  &$0.0657$ &$0.0000$ &$0.0001$  &$0.0002$ \\
\hline
\end{tabular}
\caption{Log score $\mathcal L$}
\end{subtable}
\caption{\textbf{Average AUC and log score of large firms.} The table shows (a) average AUC and (b) average log score, where $p_{\text{aud}+\text{man}}$, $p_{\text{aud}}$, and $p_{\text{man}}$ denote $p$-values from a paired two-tailed $t$-test between the scores of the current model and the three models including text. Averages and standard errors are calculated based on 10 folds, which are constructed by sampling firms larger than 5 million DKK.}\label{table:CV_performance_large}
\end{table}

\subsection{Robustness: Sampling Across Time}
In order to ensure that the observed signal in the text is not merely a result of a particular fold composition where we accidentally gauge a proxy for a temporal effect, we also perform a robustness test where we explicitly construct folds based on the publication year of the annual reports. This gives four folds in total. The results of this experiment are summarized in Table \ref{table:CV_performance_yr}, and the scores display the same tendency as in Table \ref{table:CV_performance}, further validating the results. The slightly larger $p$-values can be attributed the smaller number of folds in this experiment, resulting in a weaker statistical test. 
\\
\begin{table}
\centering
\begin{subtable}{\textwidth}
\centering
\begin{tabular}{lcccc}
\hline
Model &$\langle \text{AUC}\rangle$ &$p_{\text{aud}+\text{man}}$ &$p_{\text{aud}}$ &$p_{\text{man}}$\\
\hline
$\text{NN}_{\text{aud}+\text{man}}$    &$0.843$ &--      &--       &--      \\
$\text{NN}_{\text{aud}}$               &$0.842$ &$0.299$ &--       &--      \\
$\text{NN}_{\text{man}}$               &$0.830$ &$0.003$ &$0.014$  &--      \\
\hline
XGB                                    &$0.826$ &$0.001$ &$0.006$  &$0.175$ \\
$\text{NN}$                            &$0.822$ &$0.001$ &$0.004$  &$0.054$ \\
Logit                                  &$0.814$ &$0.000$ &$0.001$  &$0.009$ \\
\hline
\end{tabular}
\caption{AUC}
\end{subtable}%
\\
\bigskip
\begin{subtable}{\textwidth}
\centering
\begin{tabular}{lcccc}
\hline
Model &$\langle \mathcal L \rangle$ &$p_{\text{aud}+\text{man}}$ &$p_{\text{aud}}$ &$p_{\text{man}}$\\
\hline
$\text{NN}_{\text{aud}+\text{man}}$    &$0.1090$ &--       &--        &--       \\
$\text{NN}_{\text{aud}}$               &$0.1095$ &$0.4130$ &--        &--       \\
$\text{NN}_{\text{man}}$               &$0.1114$ &$0.0312$ &$0.1289$  &--       \\
\hline
XGB                                    &$0.1109$ &$0.0128$ &$0.0627$  &$0.3484$ \\
$\text{NN}$                            &$0.1122$ &$0.0112$ &$0.0166$  &$0.2098$ \\
Logit                                  &$0.1127$ &$0.0081$ &$0.0005$  &$0.1892$ \\
\hline
\end{tabular}
\caption{Log score $\mathcal L$}
\end{subtable}
\caption{\textbf{Average AUC and log score obtained from sampling years.} The table shows (a) average AUC and (b) average log score, where $p_{\text{aud}+\text{man}}$, $p_{\text{aud}}$, and $p_{\text{man}}$ denote $p$-values from a paired two-tailed $t$-test between the scores of the current model and the three models including text. Averages and standard errors are calculated based on 4 folds, which are constructed by sampling publication years.}\label{table:CV_performance_yr}
\end{table}

\section{Cases of Blocks with High Attention Weights}\label{sec:case_study_ENG}
The attention weights can be extracted from the individual word blocks to highlight words and phrases, which are important for the prediction. In this section we present examples of individual blocks from five auditors' reports. The attention-color-strength of individual words is relative to the largest weight in that particular block, and stop words are inserted for completeness to make the text more readable. In the five cases below the attention mechanism successfully highlights sections that intuitively should affect the distress prediction, e.g. ``\textit{no realistic options for obtaining funding}'' and ``\textit{significant uncertainty about the firm's ability to continue operations}''. Generally, these examples show cases of information that would be very difficult to represent in traditional features, such as those described in Table \ref{table:covariates}. The texts are originally in Danish, and we note that the translation has required shifting some words, but to the best of our ability we have aimed at a 1:1 comparison. The original texts are in \ref{app:blocks} for reference.

\subsubsection*{Example 1}
\begin{flushleft}
\fboxsep=0pt\relax
\bcolorbox{red!22.69}{henceforth. }%
\bcolorbox{red!2.25}{It }%
\bcolorbox{red!2.25}{is }%
\bcolorbox{red!2.25}{our }%
\bcolorbox{red!100.0}{assessment }%
\bcolorbox{red!2.25}{that }%
\bcolorbox{red!2.25}{there }%
\bcolorbox{red!2.25}{are }%
\bcolorbox{red!100.0}{no }%
\bcolorbox{red!100.0}{realistic }%
\bcolorbox{red!100.0}{options }%
\bcolorbox{red!2.25}{for }%
\bcolorbox{red!100.0}{obtaining }%
\bcolorbox{red!100.0}{funding }%
\bcolorbox{red!2.25}{and }%
\bcolorbox{red!2.25}{we }%
\bcolorbox{red!100.0}{therefore }%
\bcolorbox{red!100.0}{make }%
\bcolorbox{red!2.25}{the }%
\bcolorbox{red!100.0}{caveat }%
\bcolorbox{red!2.25}{that }%
\bcolorbox{red!2.25}{the }%
\bcolorbox{red!100.0}{statement }%
\bcolorbox{red!2.25}{has }%
\bcolorbox{red!2.25}{been }%
\bcolorbox{red!100.0}{submitted }%
\bcolorbox{red!100.0}{on }%
\bcolorbox{red!2.25}{the }%
\bcolorbox{red!100.0}{basis }%
\bcolorbox{red!2.25}{of }%
\bcolorbox{red!100.0}{continued }%
\bcolorbox{red!100.0}{operations. }%
\bcolorbox{red!2.25}{It }%
\bcolorbox{red!2.25}{is }%
\bcolorbox{red!2.25}{our }%
\bcolorbox{red!100.0}{opinion }%
\bcolorbox{red!2.25}{that }%
\bcolorbox{red!2.25}{the }%
\bcolorbox{red!100.0}{statement }%
\bcolorbox{red!2.25}{as }%
\bcolorbox{red!2.25}{a }%
\bcolorbox{red!100.0}{consequence }%
\bcolorbox{red!2.25}{of }%
\bcolorbox{red!2.25}{the }%
\bcolorbox{red!100.0}{significance}%
\end{flushleft}

\subsubsection*{Example 2}
\begin{flushleft}
\fboxsep=0pt\relax
\bcolorbox{red!0.96}{the }%
\bcolorbox{red!100.0}{mention }%
\bcolorbox{red!0.96}{in }%
\bcolorbox{red!0.96}{the }%
\bcolorbox{red!100.0}{statement's }%
\bcolorbox{red!100.0}{notes }%
\bcolorbox{red!0.96}{and }%
\bcolorbox{red!0.96}{the }%
\bcolorbox{red!100.0}{management's }%
\bcolorbox{red!100.0}{report }%
\bcolorbox{red!100.0}{where }%
\bcolorbox{red!0.96}{the }%
\bcolorbox{red!100.0}{management }%
\bcolorbox{red!100.0}{explains }%
\bcolorbox{red!0.96}{the }%
\bcolorbox{red!100.0}{significant }%
\bcolorbox{red!100.0}{uncertainty }%
\bcolorbox{red!0.96}{about }%
\bcolorbox{red!0.96}{the }%
\bcolorbox{red!100.0}{firm's }%
\bcolorbox{red!100.0}{ability }%
\bcolorbox{red!0.96}{to }%
\bcolorbox{red!33.57}{continue }%
\bcolorbox{red!33.57}{operations }%
\bcolorbox{red!0.96}{as }%
\bcolorbox{red!0.96}{it }%
\bcolorbox{red!0.96}{is }%
\bcolorbox{red!33.57}{still }%
\bcolorbox{red!33.57}{uncertain }%
\bcolorbox{red!0.96}{if }%
\bcolorbox{red!0.96}{the }%
\bcolorbox{red!33.57}{necessary }%
\bcolorbox{red!33.57}{liquidity }%
\bcolorbox{red!33.57}{can }%
\bcolorbox{red!33.57}{be }%
\bcolorbox{red!33.57}{generated }%
\bcolorbox{red!0.96}{for }%
\bcolorbox{red!33.57}{financing }%
\end{flushleft}

\subsubsection*{Example 3} 
\begin{flushleft}
\fboxsep=0pt\relax
\bcolorbox{red!26.27}{obtaining }%
\bcolorbox{red!2.09}{the }%
\bcolorbox{red!72.11}{liquidity }%
\bcolorbox{red!2.09}{for }%
\bcolorbox{red!72.11}{payment }%
\bcolorbox{red!2.09}{of }%
\bcolorbox{red!2.09}{a }%
\bcolorbox{red!72.11}{significant }%
\bcolorbox{red!72.11}{tax }%
\bcolorbox{red!72.11}{liability. }%
\bcolorbox{red!2.09}{It }%
\bcolorbox{red!2.09}{is }%
\bcolorbox{red!72.11}{uncertain }%
\bcolorbox{red!72.11}{whether }%
\bcolorbox{red!2.09}{the }%
\bcolorbox{red!72.11}{firm }%
\bcolorbox{red!2.09}{will }%
\bcolorbox{red!72.11}{be }%
\bcolorbox{red!72.11}{able }%
\bcolorbox{red!2.09}{to }%
\bcolorbox{red!72.11}{obtain }%
\bcolorbox{red!2.09}{this }%
\bcolorbox{red!72.11}{additional }%
\bcolorbox{red!72.11}{liquidity. }%
\bcolorbox{red!2.09}{We }%
\bcolorbox{red!2.09}{are }%
\bcolorbox{red!72.11}{thus }%
\bcolorbox{red!72.11}{not }%
\bcolorbox{red!72.11}{able }%
\bcolorbox{red!2.09}{to }%
\bcolorbox{red!72.11}{comment }%
\bcolorbox{red!2.09}{on }%
\bcolorbox{red!2.09}{the }%
\bcolorbox{red!72.11}{company's }%
\bcolorbox{red!72.11}{ability }%
\bcolorbox{red!2.09}{to }%
\bcolorbox{red!72.11}{continue }%
\bcolorbox{red!72.11}{operations }%
\bcolorbox{red!2.09}{the }%
\bcolorbox{red!49.91}{coming }%
\bcolorbox{red!49.91}{year, }%
\bcolorbox{red!49.91}{why }%
\bcolorbox{red!2.09}{we }%
\bcolorbox{red!49.91}{have }%
\bcolorbox{red!49.91}{reservations. }%
\bcolorbox{red!2.09}{It }%
\bcolorbox{red!2.09}{should }%
\bcolorbox{red!49.91}{also }%
\bcolorbox{red!49.91}{be }%
\bcolorbox{red!49.91}{noted }%
\bcolorbox{red!2.09}{that }%
\bcolorbox{red!2.09}{there }%
\bcolorbox{red!2.09}{is }%
\bcolorbox{red!49.91}{not }%
\end{flushleft}

\subsubsection*{Example 4}
\begin{flushleft}
\fboxsep=0pt\relax
\bcolorbox{red!2.91}{has }%
\bcolorbox{red!60.11}{not }%
\bcolorbox{red!33.74}{yet }%
\bcolorbox{red!60.11}{received }%
\bcolorbox{red!60.11}{acknowledgment }%
\bcolorbox{red!2.91}{from }%
\bcolorbox{red!2.91}{the }%
\bcolorbox{red!60.11}{involved }%
\bcolorbox{red!60.11}{bank, }%
\bcolorbox{red!2.91}{and }%
\bcolorbox{red!2.91}{on }%
\bcolorbox{red!2.91}{that }%
\bcolorbox{red!60.11}{basis }%
\bcolorbox{red!2.91}{we }%
\bcolorbox{red!60.11}{can }%
\bcolorbox{red!60.11}{not }%
\bcolorbox{red!60.11}{reach }%
\bcolorbox{red!2.91}{a }%
\bcolorbox{red!85.85}{conclusion }%
\bcolorbox{red!85.85}{regarding }%
\bcolorbox{red!2.91}{the }%
\bcolorbox{red!85.85}{firm's }%
\bcolorbox{red!85.85}{ability }%
\bcolorbox{red!2.91}{to }%
\bcolorbox{red!85.85}{continue }%
\bcolorbox{red!85.85}{operation. }%
\bcolorbox{red!85.85}{Non-Conclusion }%
\bcolorbox{red!2.91}{Due }%
\bcolorbox{red!2.91}{to }%
\end{flushleft}

\subsubsection*{Example 5}
\begin{flushleft}
\fboxsep=0pt\relax
\bcolorbox{red!2.25}{on }%
\bcolorbox{red!66.43}{note }%
\bcolorbox{red!66.43}{xxnumberxx }%
\bcolorbox{red!66.43}{external }%
\bcolorbox{red!66.43}{accounts, }%
\bcolorbox{red!66.43}{which }%
\bcolorbox{red!66.43}{shows }%
\bcolorbox{red!2.25}{that }%
\bcolorbox{red!2.25}{the }%
\bcolorbox{red!66.43}{company's }%
\bcolorbox{red!66.43}{equity }%
\bcolorbox{red!2.25}{is }%
\bcolorbox{red!66.43}{exhausted. }%
\bcolorbox{red!2.25}{The }%
\bcolorbox{red!66.43}{company's }%
\bcolorbox{red!66.43}{continued }%
\bcolorbox{red!60.64}{operation }%
\bcolorbox{red!60.64}{therefore }%
\bcolorbox{red!60.64}{depends }%
\bcolorbox{red!2.25}{on }%
\bcolorbox{red!2.25}{that }%
\bcolorbox{red!2.25}{the }%
\bcolorbox{red!60.64}{necessary }%
\bcolorbox{red!60.64}{liquidity }%
\bcolorbox{red!60.64}{continues }%
\bcolorbox{red!2.25}{to }%
\bcolorbox{red!2.25}{be }%
\bcolorbox{red!60.64}{provided. }%
\bcolorbox{red!2.25}{The }%
\bcolorbox{red!60.64}{firm's }%
\end{flushleft}
\noindent
\vspace{-5mm}
\begin{figure}[h]
\resizebox{\linewidth}{!}{
\begin{tikzpicture}
\begin{axis}[
    tick label style={font=\footnotesize},
    hide axis,
    scale only axis,
    height=0pt,
    width=0pt,
    colormap={whitered}{color=(white) color=(red)},
    colorbar horizontal,
    point meta min=0,
    point meta max=100,
    colorbar style={
        width=7cm,
        xtick={0, 20, 40, ..., 100}
    }]
    \addplot [draw=none] coordinates {(0,0)};
\end{axis}
\end{tikzpicture}
}
\end{figure}

\section{Outlook and Conclusion}\label{sec:conclusion}
We have introduced a network architecture consisting of both convolutional and recurrent neural networks for predicting corporate distresses using auditors' reports and managements' statements of annual reports. By concatenating the neural network model with numerical financial variables, we found that the model with auditors' reports increased the AUC by almost 2 percentage points compared to a neural network without text while the managements' statements only gave an enhancement of roughly 1 percentage point. The enhancement in model performance is statistically significant at the 1\% level in both cases, demonstrating that there is useful information to be extracted from text segments besides what is already contained in the numerical financial variables. Statistical tests also revealed that auditors' reports provide significantly more information than managements' statements and that all useful information contained in the managements' statements is contained in the auditors' reports as well. These findings suggest that further analyses should focus merely on auditors' reports. For firms with a size greater than 5 million DKK the auditors' reports enhanced the AUC by more than 2.5 percentage points, while it was still roughly 1 percentage point for managements' statements, showing that textual analysis of auditors' reports is especially useful for large firms from which we benefit the most from accurate distress predictions. 

In future work it would be interesting to investigate individual paragraphs within the auditors' reports and managements' statements to see if certain paragraphs in combination are more suited for the distress prediction.

\section*{Acknowledgements}
The authors are grateful to Benjamin Christoffersen for helpful comments. Casper Hansen and Christian Hansen are financially supported by the Innovation Fund Denmark through the DABAI project.

\clearpage 

\appendix

\section{Results of Parameter Tuning in the Main Model}
\setcounter{figure}{0} 

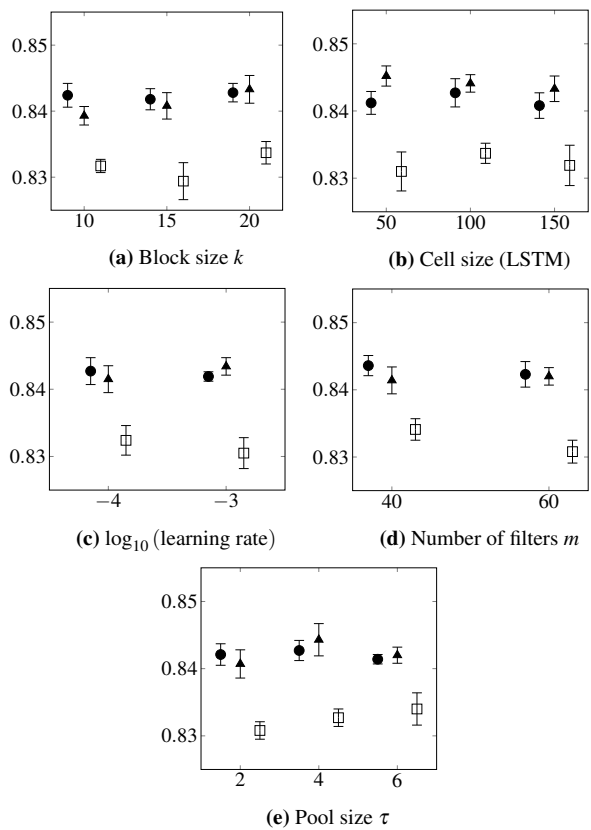
\begin{figure}[h]
\captionsetup[subfigure]{oneside,margin={0.8cm,0cm}}
\centering
\begin{subfigure}{.47\linewidth}
\resizebox{\linewidth}{!}{
\begin{tikzpicture}

\tikzstyle{every node}=[font=\huge]

\pgfplotsset{
    scale only axis,
    xmin=8, xmax=22,
    xtick={10, 15, 20},
        y tick label style={
        /pgf/number format/.cd,
            fixed,
            fixed zerofill,
            precision=2,
        /tikz/.cd,
        ytick={0.83, 0.84, ..., 0.85}}
    }

\begin{axis}[
  ymin=0.825, ymax=0.855,
  scatter/classes={a={}}]

    \addplot [scatter, only marks, mark=*,mark size=5pt]
    plot [error bars/.cd, y dir = both, y explicit]
    table[row sep=crcr, x index=0, y index=1, y error index=2,]{
     9.0 0.8424 0.0018\\
    14.0 0.8418 0.0016\\
    19.0 0.8428 0.0014\\
    };    

    \addplot [scatter, only marks, mark=triangle*,mark size=5pt]
    plot [error bars/.cd, y dir = both, y explicit]
    table[row sep=crcr, x index=0, y index=1, y error index=2,]{
    10.0 0.8393 0.0014\\
    15.0 0.8408 0.0020\\
    20.0 0.8433 0.0021\\
    };    
    
    \addplot [scatter, only marks, mark=square,mark size=5pt]
    plot [error bars/.cd, y dir = both, y explicit]
    table[row sep=crcr, x index=0, y index=1, y error index=2,]{
    11.0 0.8317 0.0010\\
    16.0 0.8294 0.0028\\
    21.0 0.8337 0.0017\\
    };

\end{axis}

\end{tikzpicture}
}
\caption{Block size $k$}
\label{fig:blocksize}
\end{subfigure}%
\,\,\,\,
\begin{subfigure}{.47\linewidth}
\resizebox{\linewidth}{!}{\begin{tikzpicture}

\tikzstyle{every node}=[font=\huge]

\pgfplotsset{
    scale only axis,
    xmin=30, xmax=170,
    xtick={50, 100, 150},
        y tick label style={
        /pgf/number format/.cd,
            fixed,
            fixed zerofill,
            precision=2,
        /tikz/.cd,
        ytick={0.83, 0.84, ..., 0.85}}
    }

\begin{axis}[
  ymin=0.825, ymax=0.855,
  scatter/classes={a={}}]

    \addplot [scatter, only marks, mark=*,mark size=5pt]
    plot [error bars/.cd, y dir = both, y explicit]
    table[row sep=crcr, x index=0, y index=1, y error index=2,]{
    41  0.8412 0.0017\\
    91  0.8427 0.0021\\
    141 0.8408 0.0019\\
    };

    \addplot [scatter, only marks, mark=triangle*,mark size=5pt]
    plot [error bars/.cd, y dir = both, y explicit]
    table[row sep=crcr, x index=0, y index=1, y error index=2,]{
    50  0.8452 0.0015\\
    100 0.8441 0.0013\\
    150 0.8433 0.0019\\
    };
    
    \addplot [scatter, only marks, mark=square,mark size=5pt]
    plot [error bars/.cd, y dir = both, y explicit]
    table[row sep=crcr, x index=0, y index=1, y error index=2,]{
    59  0.8310 0.0029\\
    109 0.8337 0.0015\\
    159 0.8319 0.0030\\
    };    
    
\end{axis}

\end{tikzpicture}}
\caption{Cell size (LSTM)}
\label{fig:cell_size}
\end{subfigure}\\[1ex]
\begin{subfigure}{.47\linewidth}
\centering
\resizebox{\linewidth}{!}{\begin{tikzpicture}

\tikzstyle{every node}=[font=\huge]

\pgfplotsset{
    scale only axis,
    xmin=-4.5, xmax=-2.5,
    xtick={-4, -3},
        y tick label style={
        /pgf/number format/.cd,
            fixed,
            fixed zerofill,
            precision=2,
        /tikz/.cd,
        ytick={0.83, 0.84, ..., 0.85}}
    }

\begin{axis}[
  ymin=0.825, ymax=0.855,
  scatter/classes={a={}}]
  
    \addplot [scatter, only marks, mark=*,mark size=5pt]
    plot [error bars/.cd, y dir = both, y explicit]
    table[row sep=crcr, x index=0, y index=1, y error index=2,]{
    -4.15 0.8427 0.0020\\
    -3.15 0.8419 0.0007\\
    };

    \addplot [scatter, only marks, mark=triangle*,mark size=5pt]
    plot [error bars/.cd, y dir = both, y explicit]
    table[row sep=crcr, x index=0, y index=1, y error index=2,]{
    -4.0 0.8415 0.0020\\
    -3.0 0.8434 0.0013\\
    };
    
    \addplot [scatter, only marks, mark=square,mark size=5pt]
    plot [error bars/.cd, y dir = both, y explicit]
    table[row sep=crcr, x index=0, y index=1, y error index=2,]{
    -3.85 0.8324 0.0022\\
    -2.85 0.8305 0.0023\\
    };    
\end{axis}

\end{tikzpicture}}
\caption{$\log_{10}{(\text{learning rate})}$}
\label{fig:eta}
\end{subfigure}%
\,\,\,\,
\begin{subfigure}{.47\linewidth}
\centering
\resizebox{\linewidth}{!}{\begin{tikzpicture}

\tikzstyle{every node}=[font=\huge]

\pgfplotsset{
    scale only axis,
    xmin=35, xmax=65,
    xtick={40, 60},
        y tick label style={
        /pgf/number format/.cd,
            fixed,
            fixed zerofill,
            precision=2,
        /tikz/.cd,
        ytick={0.83, 0.84, ..., 0.85}}
    }

\begin{axis}[
  ymin=0.825, ymax=0.855,
  scatter/classes={a={}}]

    \addplot [scatter, only marks, mark=*,mark size=5pt]
    plot [error bars/.cd, y dir = both, y explicit]
    table[row sep=crcr, x index=0, y index=1, y error index=2,]{
    37 0.8436 0.0015\\
    57 0.8423 0.0019\\
    };

    \addplot [scatter, only marks, mark=triangle*,mark size=5pt]
    plot [error bars/.cd, y dir = both, y explicit]
    table[row sep=crcr, x index=0, y index=1, y error index=2,]{
    40 0.8414 0.0020\\
    60 0.8420 0.0013\\
    };
    
    \addplot [scatter, only marks, mark=square,mark size=5pt]
    plot [error bars/.cd, y dir = both, y explicit]
    table[row sep=crcr, x index=0, y index=1, y error index=2,]{
    43 0.8341 0.0016\\
    63 0.8308 0.0017\\
    };
\end{axis}

\end{tikzpicture}}
\caption{Number of filters $m$}
\label{fig:out_channel}
\end{subfigure}\\[1ex]
\begin{subfigure}{0.47\linewidth}
\centering
\resizebox{\linewidth}{!}{\begin{tikzpicture}

\tikzstyle{every node}=[font=\huge]

\pgfplotsset{
    scale only axis,
    xmin=1, xmax=7,
    xtick={2, 4, 6},
        y tick label style={
        /pgf/number format/.cd,
            fixed,
            fixed zerofill,
            precision=2,
        /tikz/.cd,
        ytick={0.83, 0.84, ..., 0.85}}
    }

\begin{axis}[
  ymin=0.825, ymax=0.855,
  scatter/classes={a={}}]

    \addplot [scatter, only marks, mark=*,mark size=5pt]
    plot [error bars/.cd, y dir = both, y explicit]
    table[row sep=crcr, x index=0, y index=1, y error index=2,]{
    1.5 0.8421 0.0016\\
    3.5 0.8427 0.0015\\
    5.5 0.8414 0.0007\\
    };

    \addplot [scatter, only marks, mark=triangle*,mark size=5pt]
    plot [error bars/.cd, y dir = both, y explicit]
    table[row sep=crcr, x index=0, y index=1, y error index=2,]{
    2.0 0.8407 0.0021\\
    4.0 0.8443 0.0024\\
    6.0 0.8420 0.0012\\
    };
    
    \addplot [scatter, only marks, mark=square,mark size=5pt]
    plot [error bars/.cd, y dir = both, y explicit]
    table[row sep=crcr, x index=0, y index=1, y error index=2,]{
    2.5 0.8308 0.0013\\
    4.5 0.8327 0.0013\\
    6.5 0.8340 0.0024\\
    };    
\end{axis}

\end{tikzpicture}}
\caption{Pool size $\tau$}
\label{fig:gridsearch_poolsize}
\end{subfigure}%
\caption{\textbf{Parameter-tuning of the neural network.} The figure illustrates AUC (\protect\audmanMarker \,\, $\text{NN}_\text{aud+man}$; \, \protect\aucMarker \,\, $\text{NN}_\text{aud}$; \, \protect\logscoreMarker \,\, $\text{NN}_\text{man}$) for different parameter choices. Error bars denote one standard error.}
\label{fig:gridsearch}
\end{figure}

\section{Cases of Blocks with High Attention Weights (Original)}\label{app:blocks}

\subsubsection*{Example 1}
\begin{flushleft}
\fboxsep=0pt\relax
\bcolorbox{red!22.69}{fremover. }%
\bcolorbox{red!2.25}{Det }%
\bcolorbox{red!2.25}{er }%
\bcolorbox{red!2.25}{vores }%
\bcolorbox{red!100.0}{vurdering, }%
\bcolorbox{red!2.25}{at }%
\bcolorbox{red!2.25}{der }%
\bcolorbox{red!100.0}{ikke }%
\bcolorbox{red!2.25}{er }%
\bcolorbox{red!100.0}{realistiske }%
\bcolorbox{red!100.0}{muligheder }%
\bcolorbox{red!2.25}{for }%
\bcolorbox{red!2.25}{at }%
\bcolorbox{red!100.0}{fremskaffe }%
\bcolorbox{red!100.0}{finansiering }%
\bcolorbox{red!2.25}{og }%
\bcolorbox{red!2.25}{vi }%
\bcolorbox{red!100.0}{tager }%
\bcolorbox{red!100.0}{derfor }%
\bcolorbox{red!100.0}{forbehold }%
\bcolorbox{red!2.25}{for, }%
\bcolorbox{red!2.25}{at }%
\bcolorbox{red!100.0}{årsregnskabet }%
\bcolorbox{red!2.25}{er }%
\bcolorbox{red!100.0}{aflagt }%
\bcolorbox{red!100.0}{under }%
\bcolorbox{red!100.0}{forudsætning }%
\bcolorbox{red!2.25}{af }%
\bcolorbox{red!100.0}{fortsat }%
\bcolorbox{red!100.0}{drift. }%
\bcolorbox{red!2.25}{Det }%
\bcolorbox{red!2.25}{er }%
\bcolorbox{red!2.25}{vores }%
\bcolorbox{red!100.0}{opfattelse, }%
\bcolorbox{red!2.25}{at }%
\bcolorbox{red!100.0}{årsregnskabet, }%
\bcolorbox{red!2.25}{som }%
\bcolorbox{red!100.0}{følge }%
\bcolorbox{red!2.25}{af }%
\bcolorbox{red!100.0}{betydeligheden }%
\end{flushleft}

\subsubsection*{Example 2}
\begin{flushleft}
\fboxsep=0pt\relax
\bcolorbox{red!0.96}{til }%
\bcolorbox{red!100.0}{omtale }%
\bcolorbox{red!0.96}{i }%
\bcolorbox{red!100.0}{årsregnskabets }%
\bcolorbox{red!100.0}{noter }%
\bcolorbox{red!0.96}{og }%
\bcolorbox{red!100.0}{ledelsesberetningen, }%
\bcolorbox{red!100.0}{hvori }%
\bcolorbox{red!100.0}{ledelsen }%
\bcolorbox{red!100.0}{redegør }%
\bcolorbox{red!0.96}{for }%
\bcolorbox{red!100.0}{væsentlig }%
\bcolorbox{red!100.0}{usikkerhed }%
\bcolorbox{red!0.96}{om }%
\bcolorbox{red!100.0}{selskabets }%
\bcolorbox{red!100.0}{evne }%
\bcolorbox{red!0.96}{til }%
\bcolorbox{red!0.96}{at }%
\bcolorbox{red!33.57}{fortsætte }%
\bcolorbox{red!33.57}{driften, }%
\bcolorbox{red!0.96}{da }%
\bcolorbox{red!0.96}{det }%
\bcolorbox{red!33.57}{endnu }%
\bcolorbox{red!0.96}{er }%
\bcolorbox{red!33.57}{usikkert, }%
\bcolorbox{red!0.96}{om }%
\bcolorbox{red!0.96}{den }%
\bcolorbox{red!33.57}{nødvendige }%
\bcolorbox{red!33.57}{likviditet }%
\bcolorbox{red!33.57}{kan }%
\bcolorbox{red!33.57}{frembringes }%
\bcolorbox{red!0.96}{til }%
\bcolorbox{red!33.57}{finansiering }%
\end{flushleft}

\subsubsection*{Example 3} 
\begin{flushleft}
\fboxsep=0pt\relax
\bcolorbox{red!26.27}{fremskaffelse }%
\bcolorbox{red!2.09}{af }%
\bcolorbox{red!72.11}{likviditeten }%
\bcolorbox{red!2.09}{til }%
\bcolorbox{red!72.11}{betaling }%
\bcolorbox{red!2.09}{af }%
\bcolorbox{red!2.09}{en }%
\bcolorbox{red!72.11}{væsentlig }%
\bcolorbox{red!72.11}{momsgæld. }%
\bcolorbox{red!2.09}{Det }%
\bcolorbox{red!2.09}{er }%
\bcolorbox{red!72.11}{usikkert }%
\bcolorbox{red!72.11}{hvorvidt }%
\bcolorbox{red!72.11}{selskabet }%
\bcolorbox{red!2.09}{vil }%
\bcolorbox{red!72.11}{være }%
\bcolorbox{red!2.09}{i }%
\bcolorbox{red!72.11}{stand }%
\bcolorbox{red!2.09}{til }%
\bcolorbox{red!2.09}{at }%
\bcolorbox{red!72.11}{fremskaffe }%
\bcolorbox{red!2.09}{denne }%
\bcolorbox{red!72.11}{yderligere }%
\bcolorbox{red!72.11}{likviditet. }%
\bcolorbox{red!2.09}{Vi }%
\bcolorbox{red!2.09}{er }%
\bcolorbox{red!72.11}{således }%
\bcolorbox{red!72.11}{ikke }%
\bcolorbox{red!2.09}{i }%
\bcolorbox{red!72.11}{stand }%
\bcolorbox{red!2.09}{til }%
\bcolorbox{red!2.09}{at }%
\bcolorbox{red!72.11}{udtale }%
\bcolorbox{red!2.09}{os }%
\bcolorbox{red!2.09}{om }%
\bcolorbox{red!72.11}{selskabets }%
\bcolorbox{red!72.11}{evne }%
\bcolorbox{red!2.09}{til }%
\bcolorbox{red!2.09}{at }%
\bcolorbox{red!72.11}{fortsætte }%
\bcolorbox{red!72.11}{driften }%
\bcolorbox{red!2.09}{det }%
\bcolorbox{red!49.91}{kommende }%
\bcolorbox{red!49.91}{år, }%
\bcolorbox{red!49.91}{hvorfor }%
\bcolorbox{red!2.09}{vi }%
\bcolorbox{red!49.91}{tager }%
\bcolorbox{red!49.91}{forbehold }%
\bcolorbox{red!49.91}{herfor. }%
\bcolorbox{red!2.09}{Det }%
\bcolorbox{red!2.09}{skal }%
\bcolorbox{red!49.91}{endvidere }%
\bcolorbox{red!49.91}{bemærkes, }%
\bcolorbox{red!2.09}{at }%
\bcolorbox{red!2.09}{der }%
\bcolorbox{red!49.91}{ikke }%
\end{flushleft}

\subsubsection*{Example 4}
\begin{flushleft}
\fboxsep=0pt\relax
\bcolorbox{red!2.91}{har }%
\bcolorbox{red!33.74}{endnu }%
\bcolorbox{red!60.11}{ikke }%
\bcolorbox{red!60.11}{modtaget }%
\bcolorbox{red!60.11}{tilkendegivelse }%
\bcolorbox{red!2.91}{fra }%
\bcolorbox{red!2.91}{det }%
\bcolorbox{red!60.11}{involverede }%
\bcolorbox{red!60.11}{pengeinstitut, }%
\bcolorbox{red!2.91}{og }%
\bcolorbox{red!2.91}{på }%
\bcolorbox{red!2.91}{den }%
\bcolorbox{red!60.11}{baggrund }%
\bcolorbox{red!60.11}{kan }%
\bcolorbox{red!2.91}{vi }%
\bcolorbox{red!60.11}{ikke }%
\bcolorbox{red!60.11}{nå }%
\bcolorbox{red!60.11}{frem }%
\bcolorbox{red!2.91}{til }%
\bcolorbox{red!2.91}{en }%
\bcolorbox{red!85.85}{konklusion }%
\bcolorbox{red!85.85}{vedrørende }%
\bcolorbox{red!85.85}{selskabets }%
\bcolorbox{red!85.85}{evne }%
\bcolorbox{red!2.91}{til }%
\bcolorbox{red!2.91}{at }%
\bcolorbox{red!85.85}{fortsætte }%
\bcolorbox{red!85.85}{driften. }%
\bcolorbox{red!85.85}{Manglende }%
\bcolorbox{red!85.85}{konklusion }%
\bcolorbox{red!2.91}{På }%
\bcolorbox{red!85.85}{grund }%
\end{flushleft}

\subsubsection*{Example 5}
\begin{flushleft}
\fboxsep=0pt\relax
\bcolorbox{red!2.25}{på }%
\bcolorbox{red!66.43}{note }%
\bcolorbox{red!66.43}{xxnumberxx }%
\bcolorbox{red!66.43}{eksternt }%
\bcolorbox{red!66.43}{regnskab, }%
\bcolorbox{red!66.43}{hvoraf }%
\bcolorbox{red!2.25}{det }%
\bcolorbox{red!66.43}{fremgår, }%
\bcolorbox{red!2.25}{at }%
\bcolorbox{red!66.43}{selskabet }%
\bcolorbox{red!66.43}{egenkapital }%
\bcolorbox{red!2.25}{er }%
\bcolorbox{red!66.43}{tabt. }%
\bcolorbox{red!66.43}{Selskabets }%
\bcolorbox{red!66.43}{fortsatte }%
\bcolorbox{red!60.64}{drift }%
\bcolorbox{red!2.25}{er }%
\bcolorbox{red!60.64}{derfor }%
\bcolorbox{red!60.64}{afhængig }%
\bcolorbox{red!2.25}{af }%
\bcolorbox{red!2.25}{at }%
\bcolorbox{red!2.25}{der }%
\bcolorbox{red!60.64}{fortsat }%
\bcolorbox{red!60.64}{stilles }%
\bcolorbox{red!2.25}{den }%
\bcolorbox{red!60.64}{nødvendige }%
\bcolorbox{red!60.64}{likviditet }%
\bcolorbox{red!2.25}{til }%
\bcolorbox{red!60.64}{rådighed. }%
\bcolorbox{red!60.64}{Selskabets }%
\end{flushleft}
\noindent
\vspace{-5mm}
\begin{figure}[h]
\resizebox{\linewidth}{!}{
\begin{tikzpicture}
\begin{axis}[
    tick label style={font=\footnotesize},
    hide axis,
    scale only axis,
    height=0pt,
    width=0pt,
    colormap={whitered}{color=(white) color=(red)},
    colorbar horizontal,
    point meta min=0,
    point meta max=100,
    colorbar style={
        width=7cm,
        xtick={0, 20, 40, ..., 100}
    }]
    \addplot [draw=none] coordinates {(0,0)};
\end{axis}
\end{tikzpicture}
}
\end{figure}

\bibliography{litt}

\end{document}